%% file: egpaper_for_review.tex
\documentclass[10pt,twocolumn,letterpaper]{article}

\usepackage{cvpr}
\usepackage{times}
\usepackage{epsfig}
\usepackage{graphicx}
\usepackage{amsmath}
\usepackage{amssymb}
\usepackage{kotex}
\usepackage{xcolor}
\usepackage{subcaption}

\usepackage{multirow}
\usepackage{booktabs}
\usepackage[pagebackref=true,breaklinks=true,letterpaper=true,colorlinks,bookmarks=false]{hyperref}

\cvprfinalcopy 

\def\cvprPaperID{7710} 

\ifcvprfinal\fi
\begin{document}

\title{More than just an auxiliary loss: Anti-spoofing Backbone Training via Adversarial Pseudo-depth Generation}

\author{Chang Keun Paik\\
University of Toronto\\
Canada\\
{\tt\small chad.paik@mail.utoronto.ca}
\and
Naeun Ko\\
Naver Corp.\\
South Korea\\
{\tt\small naeun.ko@navercorp.com}
\and
Youngjoon Yoo\\
Naver Corp.\\
South Korea\\
{\tt\small youngjoon.yoo@navercorp.com}
}

\maketitle

\input{0_Abstract}
\input{1_Introduction}
\input{2_RelatedWork}

\input{3_ProposedMethod}

\input{4_Experiment}
\input{5_Conclusion}

{\small
\bibliographystyle{ieee_fullname}
\bibliography{egbib}
}

\end{document}

%% file: 0_Abstract.tex
\begin{abstract}
In this paper, a new method of training pipeline is discussed to achieve significant performance on the task of anti-spoofing with RGB image. 
We explore and highlight the impact of using pseudo-depth to pre-train a network that will be used as the backbone to the final classifier. 
While the usage of pseudo-depth for anti-spoofing task is not a new idea on its own, previous endeavours utilize pseudo-depth simply as another medium to extract features for performing prediction, or as part of many auxiliary losses in aiding the training of the main classifier, normalizing the importance of pseudo-depth as just another semantic information.
Through this work, we argue that there exists a significant advantage in training the final classifier can be gained by the pre-trained generator learning to predict the corresponding pseudo-depth of a given facial image, from a Generative Adversarial Network framework.
Our experimental results indicate that our method results in a much more adaptable system that can generalize beyond intra-dataset samples, but to inter-dataset samples, which it has never seen before during training.
Quantitatively, our method approaches the baseline performance of the current state of the art anti-spoofing models with $15.8${\normalfont x} less parameters used.  Moreover, experiments showed that the introduced methodology performs well only using basic binary label without additional semantic information which indicates potential benefits of this work in industrial and application based environment where trade-off between additional labelling and resources are considered.

\end{abstract}

%% file: 1_Introduction.tex
\section{Introduction}
With the rise of systems and applications that verifies users with facial recognition systems, a reliable facial anti-spoofing is required more than ever. 
This task is necessary for systems that requires face of a user as a form of verification, as failure to perform accurate face anti-spoofing can potentially lead to catastrophic security breaches.


Continuous development of sensors have enabled facial anti-spoofing to be performed with devices such as IR camera and depth camera, which arguably has more accurate and reliable data for face anti-spoofing task. However, there are still many devices that are being manufactured without the aforementioned sensors, as well as many that were made in the past without the inclusion of these technologies that still require the ability to perform facial anti-spoofing. In order to solve these issues, face anti-spoofing has traditionally been performed with RGB images obtained from camera.

Historically, face anti-spoofing has relied upon classical machine learning methods, with more development being done in the recent years that utilize the power of CNN (Convolutional Neural Network). Still, face anti-spoofing with RGB image(s) still has massive room for improvements.
Anti-spoofing task, at its core, is a simple binary classification task that predicts whether an image, or series of image is that of a real face, or a spoof. 
Despite the simplicity of the task at its core, many endeavors have been presented in an attempt to increase the performance of networks that perform anti-spoofing task.
One of these methods is through the usage of facial depth-map, obtained from an RGB image using various depth estimation techniques. These estimated depth-map of face images will be referred to as \textbf{pseudo-depth} throughout this paper, as they do not represent the actual depth-map of the given scene, as would be measured by a depth camera.  The current de-facto way of producing these pseudo-depths are through Position map Regression Network (PRNet)\cite{feng2018joint}, which have been used by other pseudo-depth based papers \cite{atoum2017face}\cite{wang2020deep}\cite{zhang2019feathernets}\cite{zhang2020celeba}. 

Rather than only using these pseudo-depths as an auxiliary loss or as another feature, this paper goes one step forward and utilizes the pseudo-depth as a dataset to pre-train a backbone for the final classifier.
In this paper, we suggest that the generation ability of the backbone to capture the pseudo-depth, reflecting the details of the real and fake images, is an essential factor for the anti-spoofing task. 
We show that the generator network, trained to predict a pseudo-depth given an image by adversarial training \cite{goodfellow2014generative}, can be used as a backbone for the real and fake classification with no or just a few updates.
From the extent experiments from various real-world datasets, we show that our proposed method outperforms the cases using the backbone pre-trained by classification based loss, and the other methods using complicated auxiliary loss.

Our contributions are summarized as follows:
(1) We show that the pseudo-depths to the real and fake images embed core information that can discriminate the real and fake images.
(2) We propose a training method for a generator network predicting a pseudo-depth of an image and show that the generator is suitable for an anti-spoofing task without any additional semantic information.
(3) We verify that classifier trained this way has better adaptability towards data that it has not seen during training by analyzing its inter-dataset classification performance.
 (4) We introduce lightweight anti-spoofing model that achieves comparable performance to the current state of the art model with only 0.71M parameters.

%% file: 2_RelatedWork.tex
\section{Related Work}
\paragraph{Image Based Methods}Prior to the routine usage of deep learning in computer vision tasks, classical machine learning algorithms such as Histogram of Oriented Gradients (HoG) \cite{schwartz2011face}\cite{komulainen2013context}, and Local Binary Pattern (LBP) \cite{chingovska2012effectiveness}\cite{boulkenafet2015face} \cite{maatta2011face} \cite{tirunagari2015detection} to extract meaningful features from facial images to perform classification. Furthermore, researchers explored the possibility of representing the image in different way to extract rich features. Representing the original image as HSV and YCbCr\cite{boulkenafet2015face} \cite{atoum2017face}, or as frequency map with Fourier analysis \cite{da2012video}\cite{li2004live}.
With the rise of deep learning methodologies in recent years, many studies naturally turned to use CNN as the main engine behind their anti-spoofing algorithms. 
Studies like \cite{yang2014learn} features a very minimal end-to-end utilization of CNN for binary classification, where a single image is passed through multiple CNN layers, followed by fully connected layers to make final prediction. Other work such as \cite{li2016original} uses pre-trained VGG-Face model as a feature extractor in conjunction with SVM to perform binary classification.

\paragraph{Temporal Based Methods}Beyond looking at single images, there are many studies that use videos to take advantage of temporal information that is present. These methods can either observe the dynamics of face that can only be present in a live sample such as blinking of the eyes \cite{tirunagari2015detection}\cite{pan2007eyeblink}, movement of lips \cite{tirunagari2015detection}, or pulse signals\cite{li2016generalized}, or non-biometric temporal signals such as Haralick features \cite{agarwal2016face}, motion estimation \cite{siddiqui2016face} or technique shown in \cite{da2012video}, where they observed the "visual rhythm" of the video to perform anti-spoofing.
In recent years, methods such as 3D Convolution\cite{gan20173d}, or 2D CNN, conjunction with Recurrent Neural Networks (RNN) \cite{Liu_2018_CVPR} or Long Short Term Memory (LSTM) cells \cite{xu2015learning} were used to leverage the rich features embedded in temporal nature of video sequence.

\paragraph{Utilization of Depthmap as Auxiliary Feature}
Various methods of training and feature extraction has been explored to enhance the performance of facial anti-spoofing algorithms. One of the rising information that researches began to focus on is the utilization of depth-map. The first of its kind was introduced in \cite{atoum2017face}, where Fully Convolutional Layer (FCN) was trained in to generate corresponding pseudo-depth of a given image. This pseudo-depth is then used as another vessel to extract features from and aid in generating the final classification. \\
Most recently, Zhang, Et al. introduced AE-Net\cite{zhang2020celeba}, where pseudo-depth is used as another auxiliary loss along with binary label, class label, and reflection map to train the network. Similarly to \cite{atoum2017face}, AE-Net has an additional FCN connected to the main classifier to generate a pseudo-depth of a given RGB image, which is then used in conjunction with pixel-wise loss to update the weights.\\

The above methods only treat the pseudo-depth as an additional feature to use for auxiliary loss, but this paper argues that the information embedded in them is too significant to just treat it as another loss. To the best of our knowledge, this is the first method that utilizes the depth-map as a means to pre-train the weights of CNN to obtain a meaningful starting weight for training the final binary classifier and achieving significant result for anti-spoofing tasks, especially for inter-dataset setting.

 \begin{figure}[t]
     \centering
     \includegraphics[width=\linewidth]{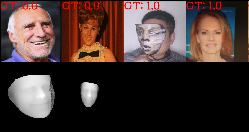}
     \caption{Pseudo-depth of live and spoof images. \textcolor{red}{Red text} on top indicates the ground truth label used for each sample}
     \label{fig:pseudodepth}
 \end{figure}

%% file: 3_ProposedMethod.tex
\section{Proposed Method}

\begin{figure*}[t]
    \centering
    \includegraphics[width=\linewidth]{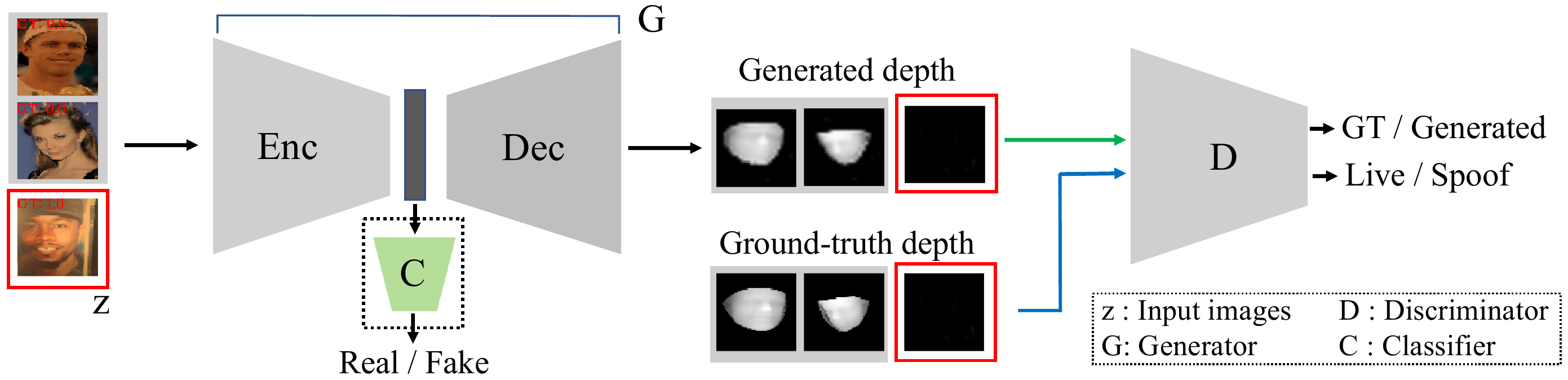}
    \vspace{-2mm}
    \caption{Architecture Illustration of Models used. Best viewed in colour}
    \vspace{-2mm}
    \label{fig:pdgan}
\end{figure*}

\begin{figure*}[t]
    \centering
    \includegraphics[width=0.95\linewidth]{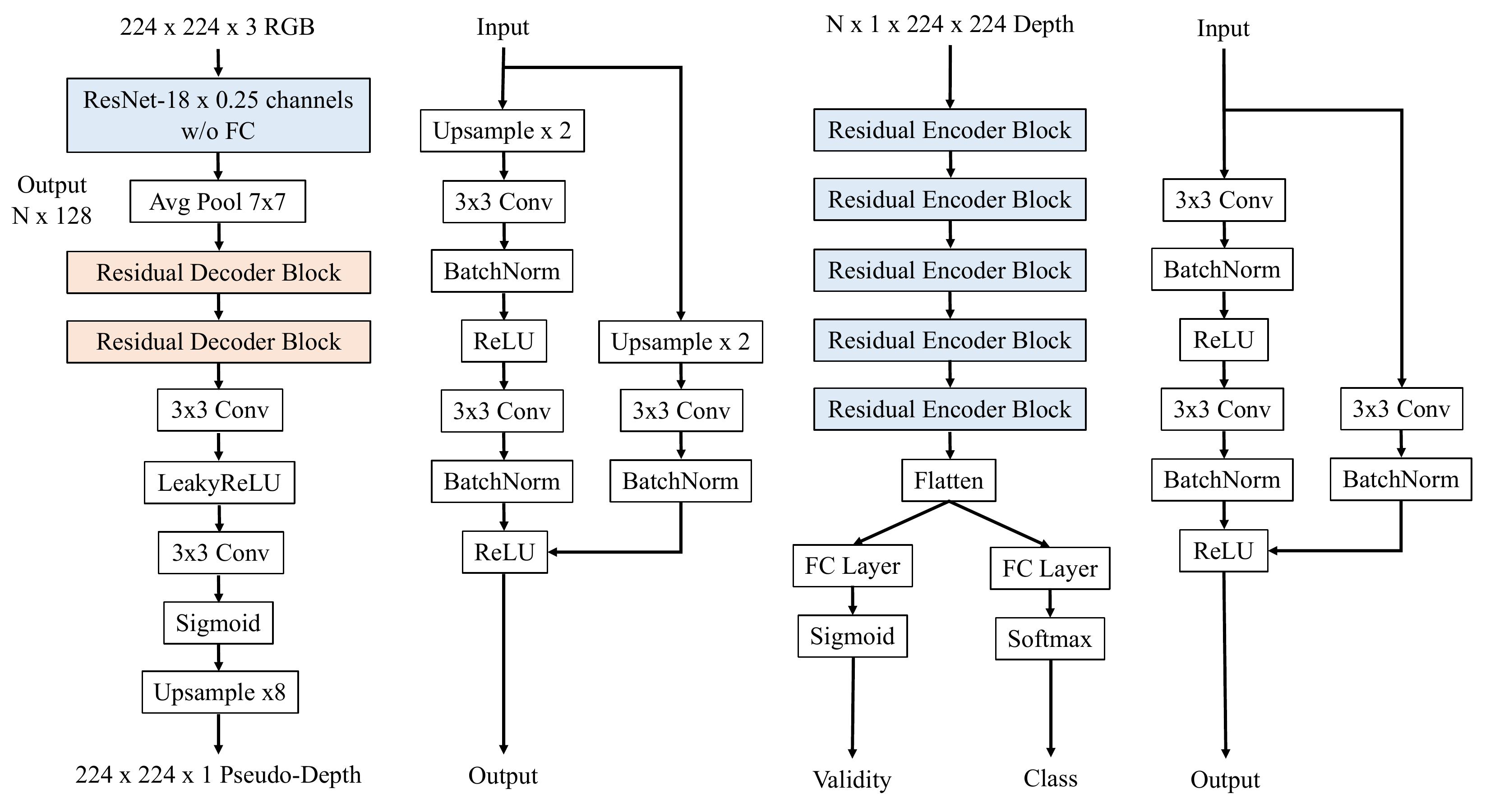}
    \vspace{-2mm}
    \caption{Illustration of the proposed network architectures. From left to right: Generator, Residual decoder block, Discriminator, and Residual encoder block.}
    \label{fig:gen_dis5}
    \vspace{-2mm}
\end{figure*}

\subsection{Pre-processing}
First, we define the pseudo-depth, as depth-map estimated by PRNet\cite{feng2018joint} for RGB images that are labelled as "live", and array initialized as zero to represent the lack of depth by RGB images that are labelled as "fake". These ground truth pseudo-depths generated and are used to train the backbone with GAN \cite{goodfellow2014generative} paradigm.
Figure \ref{fig:pseudodepth} illustrates a few examples of ground truth pseudo-depth images prepared to be used to train the backbone.

\subsection{Pseudo-Depth GAN}
\label{arch:gan}
We train the backbone network for the spoof-image classification by using a bespoke Generative Adversarial Network for this purpose, named Pseudo-Depth GAN (PDGAN).
The backbone training was done through an adversarial fashion, specifically basing the architecture and methods of Pix2Pix\cite{isola2017image}, as the task of generating corresponding pseudo-depth given an RGB image is an image-to-image translation task. 
Figure~\ref{fig:pdgan} shows the overall framework of the proposed PDGAN structure.
An encoder-decoder network was set up, where the encoder is learned to generate a proper embedding for the input image, which is then used as the input to the decoder to produce the translated image. The generator was trained with combination of GAN objective loss and $L1$ pixel loss to enhance the quality and accuracy of the output pseudo-depth.\\
However, unlike Pix2Pix\cite{isola2017image} where a Patch-Based critic is used to output a final binary classification that measure how "real" the image is, a critic with an auxiliary classifier as introduced by Odena et al.\cite{odena2017conditional} is used. This is due to the input images having clear differences in their class (live or spoof), and how the corresponding depth are drastically different. The usage of auxiliary classifier in critic aided in stabilization of the training. The architectural details of PDGAN is illustrated in Figure \ref{fig:gen_dis5}. 

The critic's objective function, $\mathcal{L}_{D}$ is a combination of two objective functions as it has the original GAN objective of distinguishing between real samples and generated (fake) samples ($\mathcal{L}_{d}$), and correctly classifying between live and spoof images ($\mathcal{L}_{c}$).
\begin{equation}
\begin{split}
\mathcal{L}_{d} &= \mathbb{E}_{x}[\text{log}(D(x))] + \mathbb{E}_{z}[\text{log}(1-D(G(z))]\\
\mathcal{L}_{cd} &= \mathbb{E}_{x,c}[\text{log}(D(x, c))] + \mathbb{E}_{z,c}[\text{log}(D(G(z), c))]\\
\mathcal{L}_{D} &= \lambda_{d}\mathcal{L}_{d} + \lambda_{c}\mathcal{L}_{c}
\end{split}
\end{equation}
where $x$ is GT depth, $z$ is RGB image input to G, and $c$ is class label of input, where $c \in \{\text{live}=0, \text{spoof}=1\}$.

Consequently, the generator's objective function, $\mathcal{L}_{G}$ is represented as the following:
\begin{equation}
\begin{split}
\label{eq:generator}
    \mathcal{L}_{G} =  &\lambda_{g}\mathbb{E}_{z}[\text{logD}(D(G(z))] \\
    + &\lambda_{cg}\mathbb{E}_{z,c}[\text{log}(D(G(z), c))] \\
    + &\lambda_{l}\mathcal{L}_{L1}(G),
\end{split}
\end{equation}
where $\mathcal{L}_{L1}(G)$ is defined as:
\begin{equation}
    \mathcal{L}_{L1}(G) = \mathbb{E}_{x,z}||x - G(z)||_{1}.
\end{equation}
The pseudo-depth corresponding to the live images has rather small variation within the data. Therefore, in order to preserve as much as detail as possible and prevent blurring, $L1$ loss is used instead of $L2$ loss.

\subsubsection{Training Settings}
The following section compiles the specific training settings of the proposed PDGAN and the classification fine-tuning. 
\paragraph{Architecture} 
Variation of ResNet-18\cite{he2016deep}, named ResNet-18x0.25 where the number of channels are reduced by 1/4 of original ResNet-18 architecture's channel, pre-trained on ImageNet\cite{imagenet_cvpr09}, was used as the backbone. The number of parameters of ResNet-18x0.25 is approximately $15.8$x less than that of original ResNet-18.
PyTorch \cite{NEURIPS2019_9015} was used to implement the models. \\

\begin{figure*}
    \centering
    \includegraphics[width=0.95\linewidth]{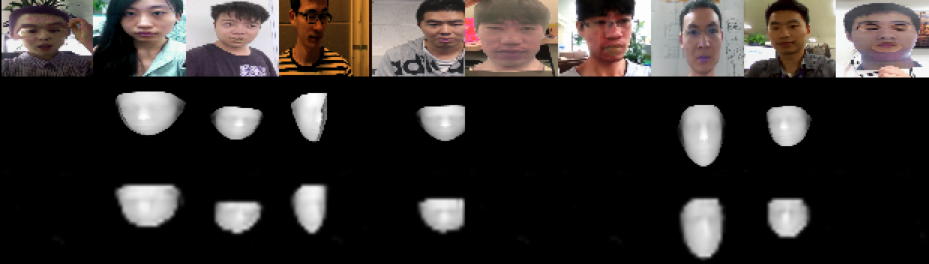}
    \vspace{-2mm}
    \caption{\textbf{Top Row}: Input RGB images from test set, \textbf{Middle Row}: GT pseudo-depth, \textbf{Bottom Row}: Generated pseudo-depth from successfully trained PDGAN.}
    \label{fig:pdgan_result}
    \vspace{-2mm}
\end{figure*}

\paragraph{Training the PDGAN}
The generator went through a "warm-up" period, where it was trained with only pixel-wise $L1$-loss for 5 epochs before entering the stage where adversarial loss from the critic is added. Learning rate of $lr=1\mathrm{e}{-4}$ was used during the warm-up period. 
Throughout the main training period and warm-up period, Adam optimizer \cite{kingma2014adam} with $lr=1\mathrm{e}{-4}$, $\beta_{1} = 0.5$, and $\beta_{2} = 0.999$ was used for the generator, while an SGD optimizer with $lr=1\mathrm{e}{-5}$ and $\text{momentum} = 0.0$ was used for the critic.
Additionally, the critic is trained every 25 steps of generator training. Though it is rare to balance training by training the critic less than the generator, the nature of the data that was being generated results in a very unstable training episodes when done using "conventional" setting. Pseudo-depth of faces generated by PRNet\cite{feng2018joint} by nature has a very small variation within the pixels, with around half of the pixels of image being values of $0$. 
Furthermore, the pseudo-depth of spoof images (regardless of attack type), which occupy more than half of the dataset distribution is a simple black image represented by matrix filled with the value of $0$. We noticed that, unless the pixel-wise loss is drastically larger than critic loss, the generator simply outputs zeros as its generated output, and critic's loss saturates. We believe that the generator abuses the fact that the data is mostly zeros, meaning it is able to lower the pixel-wise loss enough while reaching the desired critic objective. Therefore, the final setting for experiment was ran with this carefully tuned parameters: \\
$\lambda_{l} = 50 \text{ for inter-dataset}, \lambda_{l} = 100 \text{ for intra-dataset}$, $\lambda_{g} = 0.2$, $\lambda_{cg} = 0.1$, $\lambda_{d} = 1$, $\lambda_{cd} = 1$. \\
Remaining training setting that is specific for intra and inter dataset experiments will be listed in section \ref{exp:types}.

The trained PDGAN is expected to produce pseudo-depth that corresponds to the input RGB image. This is illustrated in Figure \ref{fig:pdgan_result}. For live images, it needs to correctly estimate the facial location and geometry, and produce a depth-map that is close to that of generated from PRNet\cite{feng2018joint}. Simultaneously, it needs to output zeros for spoof images.

\paragraph{Classifier fine-tuning}
After the generator is sufficiently trained, encoder of the generator is set as the backbone of the final classifier network. The embedding vector from the final generator is then flattened and passed into two hidden layers to output final liveness probability. Binary cross entropy loss is used to optimize the parameters using Adam optimizer \cite{kingma2014adam} with $lr=1\mathrm{e}{-3}$, $\beta_{1} = 0.9$ and $\beta_{2} = 0.999$.

\paragraph{Pre-processing}
Facial bounding boxes are obtained through RetinaFace\cite{deng2019retinaface} and is used to crop only the facial region from the RGB image. The cropped images are then resized into $224$x$224$, and normalized to be in range of $[0,1]$ before entering the network.\\
Furthermore, the following augmentations were applied to training samples: Geometric transformation was applied to the face bounding box of each image to give variation of facial area and shape seen by the network. GT pseudo-depths of images labelled as "live" underwent the same set of transformation (since GT pseudo-depths of images labelled "fake" is a simple uni-valued array, there was no point in applying the same geometrical transformation).  Colour augmentation, modifying brightness, hue-saturation, and temperature was applied to the input RGB image only, as we believe that the resulting depth-map should only be dependent on the geometrical aspects of an image.

%% file: 4_Experiment.tex
\section{Experiment Setting}

\subsection{Experiment Types}
\label{exp:types}
In this section, we explore the two experiments with various quantitative analysis to fully reveal the advantages that backbone pre-training with pseudo-depth brings. 
Notably, the \textbf{Intra-Dataset Experiment}, where train and test data are both from the same domain, and \textbf{Inter-Datset Experiment}, where train and test dataset are from different domains, are concerned in the experiment. 
\subsubsection{Intra-Dataset Experiment}
\label{exp:intra}
As the name suggests, fully intra-dataset experiment only uses one dataset. This means that the test data that is used to evaluate the performance of the model is that of same domain as the data used to train the model. \textbf{CelebA-Spoof} dataset\cite{zhang2020celeba} was a suitable choice for the dataset, as it is the most recently released dataset, and the largest publicly available dataset for anti-spoofing tasks, significantly. 
CelebA-Spoof dataset already separates the train and test data, and ensures that they have no crossover in facial identification to preserve the integrity of a fair test set.
\begin{figure*}
    \centering
    \includegraphics[width=0.99\linewidth]{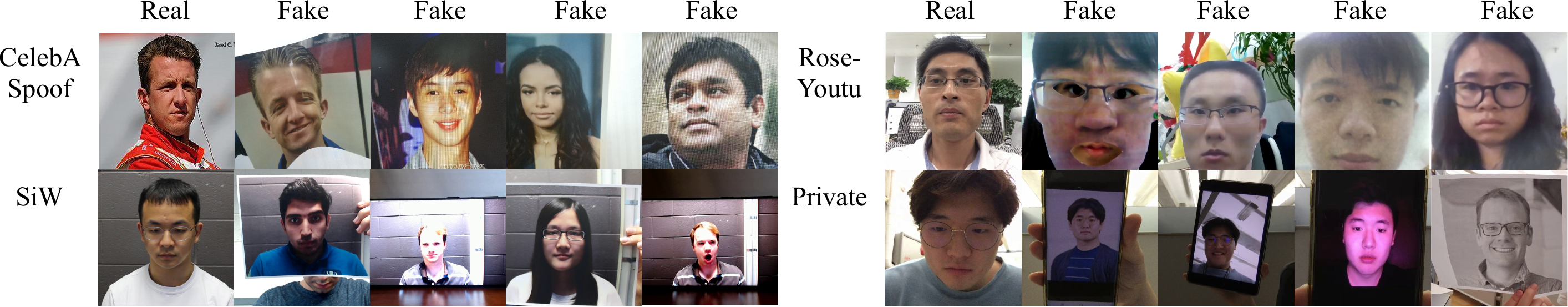}
    \vspace{-2mm}
    \caption{Examples from each training dataset}
    \label{fig:dataset}
    \vspace{-2mm}
\end{figure*}
\begin{figure*}
    \centering
    \includegraphics[width=0.99\linewidth]{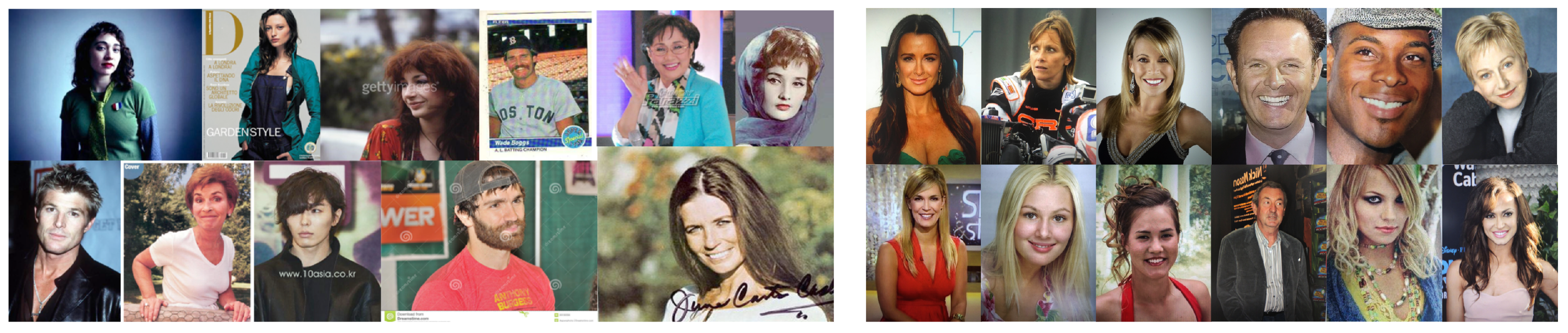}
    \vspace{-2mm}
    
    \caption{Failure cases of our model. \textbf {Left}: Real images that our model predict as spoof. \textbf{Right}: Spoof images that our model predict as real. Qualitatively, real samples have vivid features of spoof images (watermarks, text, segmented face)  whereas spoof samples are visually close to real samples.}
    \label{fig:predict_wrong}
\end{figure*}
The full train samples from CelebA-Spoof dataset also had their corresponding ground truth pseudo-depths generated to follow the training methods mentioned in Section \ref{arch:gan}.

\paragraph{Epoch Normalization}
\label{sec:epoch_norm}
During PDGAN training, the network already trains with the data that will later be used for classifier training. Though indirectly, this allows the encoder to adapt to the dataset prior to the classification, and will have an advantage over other experiments that does not use the PDGAN based backbone. Therefore, normalization of epochs was performed to adjust for this advantage by running the classifier with PDGAN based backbone smaller number of epochs than the networks that does not have prior knowledge of the dataset foregoing the classifier training session. 
Backbone training was done for 12 epoch in total (including the 5 epochs of warm-up stage), after which the encoder is taken as the backbone to the final classifier. The classifier training was done for 50 epochs in total for baseline network, and applying the aforementioned normalization, we run the network with PDGAN based backbone for 38 epochs.

\subsubsection{Inter-Dataset Experiment}
\label{exp:inter}
The main aspect of inter-dataset experiment is to see how the network trained under certain domain of data performs on another domain of data which it has not seen during training. This ability to perform well on style of samples is important as it can demonstrate the generalization property of the network. 
For training with backbone, \textbf{Rose-Youtu}~\cite{li2018unsupervised} and \textbf{Spoofing-in-the-Wild}~\cite{Liu_2018_CVPR} is used. Pseudo-depth generation was performed as in Section \ref{exp:intra}. 
We trained the classifier with the same dataset with an addition of a custom set collected, similar to the aforementioned two datasets. It consists of real samples, and spoof attacks in the forms of printed paper, mobile phone screen, and tablet screen. 
The evaluation data used for inter-dataset experiment is identical to the data used in Section \ref{exp:intra}. PDGAN was again trained for 12 epochs, and classifier training was done for 20 epochs. 
The network with PDGAN based backbone was ran for 8 epochs.

\subsection{Evaluation Metrics}
Most anti-spoofing evaluations commonly use the existing classification metrics including Bona Fide Presentation Classification Error Rate (BPCER), Attack Presentation Classification Error Rate (APCER), and Average Classification Error Rate (ACER) \cite{boulkenafet2017oulu}. 
Alongside the three de-facto methods, Area under Curve (AUC) and F1-score are used as well. The 5 metrics (BPCER, APCER, ACER, AUC, F1) are used for all three experimental settings from \ref{exp:types}. Although HTER\cite{de2013can} is a commonly used metric for inter-dataset evaluation, we used the same evaluation metrics for each intra-dataset and inter-dataset.

Consider an application where an anti-spoof dataset needs to be integrated. If there are enough data collected to train a neural network, then the samples it will encounter after deployment can be thought of as an intra-dataset case, which then the 5 metrics will be used to perform benchmark to determine the performance of the model. If there are not enough data collected to train but only to perform benchmark, then this case could be thought of as inter-dataset, but the application itself, and the data that it will be seeing are identical to that of previous case. From an application view point, we strongly believe that there is no need to separate the evaluation metric between intra and inter dataset, and believe that the 5 metrics mentioned earlier is suitable for inter-dataset as well.
For those evaluation metrics which are dependent on threshold value (ACER, APCER, BPCER, F1), threshold values from $[0,1]$ are exhaustively checked to determine which threshold achieves the best ACER. Therefore, the values of evaluation metrics other than AUC is determined using a specific threshold that will be recorded in the final result.

\section{Experiment Results}
\subsection{Intra Dataset Experiment}
\input{tables/intra_table_main}
Experiments were conducted and compared between classifier with backbone that is of ResNet-18x0.25 pre-trained with ImageNet\cite{imagenet_cvpr09} and with PDGAN encoder as backbone.
As in Table~\ref{table:intra_main}, using backbone trained with PDGAN outperforms the network with backbone pre-trained with ImageNet, even after epoch normalization is applied to ensure that each network views the data the same amount. Current model was also compared with current state-of-the-art model, AENet\cite{zhang2020celeba}. Though the raw value of performance is lower, considering how the number of parameters of our model is lower by approximately 15.8 times, the performance of our model is arguably comparable.

\subsubsection{Ablation Study: Multihead model}
Usage of additional semantic information, notably the specific labels for spoof images are used to enhance the performance of the classifier. In order to observe the effect of using semantic information, classifier was modified to have one more output for prediction of specific class as well, with same weights on both losses. This new classifier with additional output is named as \textbf{Multihead model}. The end results are compared with the original classifier that only used live/spoof binary information.

\input{tables/intra_table_ablation}
This table shows the following results: 1) Semantic information itself may not be required in every network, especially in a smaller sized network, where learning with semantic information may acting as an inhibitor to learn the main task. This is shown by juxtaposing the two ResNet based network, and seeing that the values are rather similar, but multihead performs slightly worse. Comparing with AENet\cite{zhang2020celeba} which demonstrated the importance of semantic information, our network has significantly less parameter size, which is the most notable factor for this discrepancy. Nonetheless, this result shows that semantic information may not have equal effectiveness as the binary information, which is the main objective of this network.
2) Semantic information should not be included when using PDGAN encoder. The difference in performance between the two PDGAN-based classifier is much larger than the differences that is observed between the two ResNet based networks. This is believed due to the pseudo-depth implicitly captures only binary class information, where there is some depth for live images, but is the identical uni-valued array for all spoof samples.
Therefore, the pre-trained encoder is already well fitted to binary classification information, and therefore not need additional semantic information. This fact may come as a significant advantage to many applications and industrial usage, where additional data labelling directly correlates to extra resources spent.  Though verbatim, it is important to emphasize that these results are recorded after epoch normalization from Section \ref{sec:epoch_norm}, meaning the PDGAN based backbone is not performing well because it interacted with the training data prior to the classifier training.

\subsection{Inter Dataset Experiment}
\input{tables/inter_table_main}
Inter-dataset experiment was conducted to explore the generalization ability of the classifier trained with proposed method compared to baseline models. 
As previously detailed in Section \ref{exp:inter}, the PDGAN prepared for inter dataset experiment was trained using Rose-Youtu\cite{li2018unsupervised} and Spoofing-in-the-Wild\cite{Liu_2018_CVPR} dataset. Then, actual classifier training was done with the two aforementioned datasets, as well as a custom dataset. 
As seen in Table \ref{table:inter_main}, other than APCER, classifier trained with PDGAN backbone outperforms the network using backbone pre-trained on ImageNet.  
\input{tables/inter_table_ablation}
\begin{figure*}[h!]
    \centering
    \begin{subfigure}[t]{0.32\textwidth}
        \centering
        \includegraphics[width=0.99\linewidth]{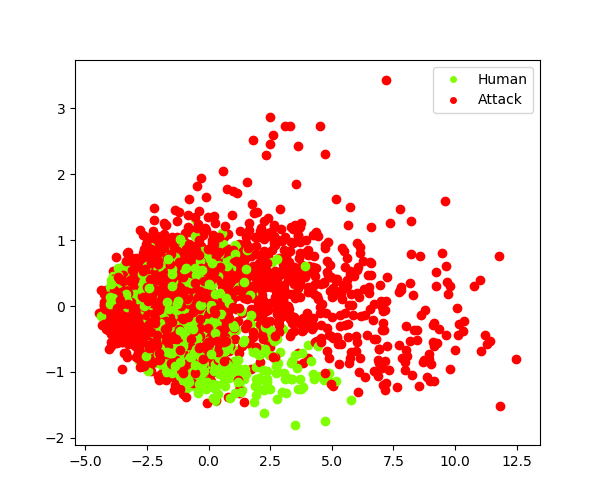}
        \caption{ResNet-18x0.25: He-initialized~\cite{he2015delving}.}
    \end{subfigure}\hfill
    ~ 
    \begin{subfigure}[t]{0.32\textwidth}
        \centering
        \includegraphics[width=0.99\linewidth]{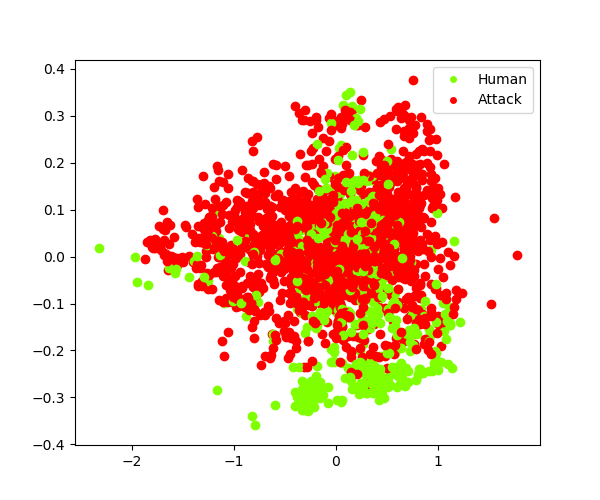}
        \caption{ResNet-18x0.25: ImageNet pre-trained.}
    \end{subfigure}\hfill
    ~ 
    \begin{subfigure}[t]{0.32\textwidth}
        \centering
        \includegraphics[width=0.99\linewidth]{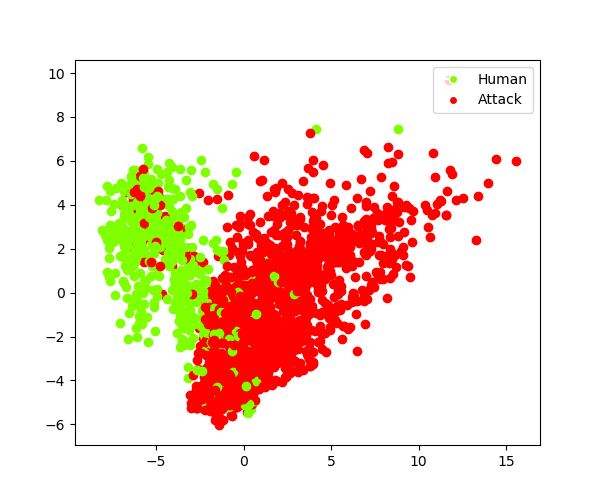}
        \caption{PDGAN Encoder}
    \end{subfigure}\hfill
    \vspace{-2mm}
    \caption{Comparison between embedding vectors from different encoders. Each plot shows the PCA projection results of (a) He-initialized, (b) ImageNet pre-trained, and (c) PDGAN pre-trained result, respectively. The result shows the ability of our embedding of discriminating the real and spoofed images. Best viewed in colour.}
    \vspace{-2mm}
    \label{fig:embedding}
\end{figure*}
\subsubsection{Ablation Study: Multihead model}
Similar in fashion as intra-dataset, ablation study was done with inter-dataset to observe the effect of extra semantic information. Again, the original binary-output classifiers were converted to their multihead counterparts and was trained with same weights for respective objectives.

The result from Figure \ref{table:inter_ablation} clearly demonstrates that using backbone trained with PDGAN outperforms using ImageNet based backbone.
The results are also similar in fashion as the intra-dataset experiment, where the addition of the extra semantic information in PDGAN backbone based classifier acts as an inhibitor, presumably due to the same reason of being implicitly trained on binary information during backbone training phase.\\
Although same epoch normalization is applied before comparing the results, inter-dataset results can be compared with bit more leniency as the data used for training, both backbone and classifier, are totally different from the evaluation dataset, meaning more interaction the network had with the dataset may not necessarily translate to better performance against inter-domain data.

\subsection{Embedding Space Visualization}
Further analysis was done to analyze the ramification of training the backbone through PDGAN. The backbone of classifier ultimately has the role of properly engineering the features given an input, which hidden layers in the form of multilayer perceptron can make final classification. The backbone weights are not frozen during the classification training hence it will continue to change and optimize, but the importance of initial weight cannot be overlooked. Therefore, the initial weights of ResNet-18x0.25 backbone pre-trained on ImageNet was compared to the encoder trained with PDGAN by visualizing the embedding it produce. Partial set from the custom dataset mentioned in Section \ref{exp:inter} is used for the embedding generation input. Since this step is testing the backbones prior to the classification training, it is guaranteed that none of these backbones tested has trained with the input data. We project the embedding into two-dimensional space by principal component analysis (PCA).  
As observed in Figure~\ref{fig:embedding}, the embedding from PDGAN's encoder shows a clear sign of discriminative capacity of the backbone trained by our method. Interestingly for this case, it can be observed that the ImageNet pre-trained model does not have much superiority over random weight initialization, which would be because that the required ability anti-spoofing and general object classification is largely different.

%% file: tables/intra_table_main.tex
\begin{table*}[t!]
\centering
\tabcolsep=0.10cm
\caption{\textbf{Evaluation on intra-dataset.} \textbf{Bold} is the best result with ResNet-18x0.25. \textcolor{blue}{\textbf{Bold}} is the current state-of-the-art. $\downarrow$ means lower value is better, and vice versa for $\uparrow$. \# Params for AENet estimated with original ResNet-18.}
\begin{tabular}[ht]{@{}lcccccccc@{}}
\toprule
\multirow{2}{*}{Model} & 
\multirow{2}{*}{\# Params (M)} &
\multirow{2}{*}{BPCER $\downarrow$} &
\multirow{2}{*}{APCER $\downarrow$} &
\multirow{2}{*}{ACER $\downarrow$} &
\multirow{2}{*}{F1 $\uparrow$} &
\multirow{2}{*}{Threshold} &
\multirow{2}{*}{AUC $\uparrow$} & \\
                                  &
                                  &                            
                                  &
                                  &                                  
                                  &        \\ 
\midrule
$\text{AENet}_{\mathcal{C}, \mathcal{S}, \mathcal{G}}$\cite{zhang2020celeba} & \textcolor{blue}{{\textbf{11.2}}}  &  \textcolor{blue}{\textbf{0.009}}  & \textcolor{blue}{\textbf{0.0229}}  & \textcolor{blue}{\textbf{0.0163}}  & -  & - & \textcolor{blue}{\textbf{0.9989}}\\ 
Resnet-18x0.25 ImageNet & 0.71  &  0.0958  & 0.0837  & 0.08975  & 0.9097  & 0.6262 & 0.9723\\
PDGAN Backbone (Ours) & 0.71  &  \textbf{0.0618}  & \textbf{0.0653}  & \textbf{0.06355} & \textbf{0.9366}  & 0.2545 & \textbf{0.9845}\\
\midrule
\end{tabular}
\vspace{-3mm}
\label{table:intra_main}
\end{table*}








%% file: tables/intra_table_ablation.tex
\begin{table*}[t!]
\centering
\tabcolsep=0.10cm
\caption{\textbf{Evaluation on intra-dataset.} \textbf{Bold} is the best result. \textcolor{blue}{\textbf{Bold}} is the current state-of-the-art. $\downarrow$ means lower value is better, and vice versa for $\uparrow$.}
\begin{tabular}[ht]{@{}lcccccccc@{}}
\toprule
\multirow{2}{*}{Model} & 
\multirow{2}{*}{\# Params (M)} &
\multirow{2}{*}{BPCER $\downarrow$} &
\multirow{2}{*}{APCER $\downarrow$} &
\multirow{2}{*}{ACER $\downarrow$} &
\multirow{2}{*}{F1 $\uparrow$} &
\multirow{2}{*}{Threshold} &
\multirow{2}{*}{AUC $\uparrow$} & \\


                                  &
                                  &                            
                                  &
                                  &                                  
                                  &        \\ 
\midrule
$\text{AENet}_{\mathcal{C}, \mathcal{S}, \mathcal{G}}$\cite{zhang2020celeba} & \textcolor{blue}{{\textbf{11.2}}}  &  \textcolor{blue}{\textbf{0.009}}  & \textcolor{blue}{\textbf{0.0229}}  & \textcolor{blue}{\textbf{0.0163}}  & -  & - & \textcolor{blue}{\textbf{0.9989}}\\ 
Resnet-18x0.25 ImageNet & 0.71  &  0.0958  & 0.0837  & 0.08975  & 0.9097  & 0.6262 & 0.9723\\
Resnet-18x0.25 ImageNet Multihead & 0.72  &  0.1008  & 0.1303  & 0.1156  & 0.8862  & 0.2554 & 0.9592\\
PDGAN Backbone (Ours) & 0.71  &  \textbf{0.0618}  & \textbf{0.0653}  & \textbf{0.0636}  & \textbf{0.9366}  & 0.2545 & \textbf{0.9845}\\
PDGAN Backbone (Ours) Multihead & 0.72  &  0.1036  & 0.0743  & 0.0890   & 0.9097  & 0.0974 & 0.9737\\
\midrule
\end{tabular}
\vspace{-3mm}
\label{table:intra_ablation}
\end{table*}


%% file: tables/inter_table_main.tex
\begin{table*}[t!]
\centering
\tabcolsep=0.10cm
\caption{\textbf{Evaluation on inter-dataset.} \textbf{Bold} is the best result. $\downarrow$ means lower value is better, and vice versa for $\uparrow$.}
\begin{tabular}[ht]{@{}lcccccccc@{}}
\toprule
\multirow{2}{*}{Model} & 
\multirow{2}{*}{\# Params (M)} &
\multirow{2}{*}{BPCER $\downarrow$} &
\multirow{2}{*}{APCER $\downarrow$} &
\multirow{2}{*}{ACER $\downarrow$} &
\multirow{2}{*}{F1 $\uparrow$} &
\multirow{2}{*}{Threshold} &
\multirow{2}{*}{AUC $\uparrow$} &\\

                                  &
                                  &                            
                                  &
                                  &                                  
                                  &        \\ 
\midrule
Resnet-18x0.25 ImageNet & 0.71  &  0.4229  & \textbf{0.2360}  & 0.3295  & 0.6366  & 0.9847 & 0.7294\\
PDGAN Backbone (Ours) & 0.71  &  \textbf{0.3987}  & 0.2469  & \textbf{0.3228}   & \textbf{0.6507}  & 0.9859 & \textbf{0.7326}\\
\midrule
\end{tabular}
\vspace{-3mm}
\label{table:inter_main}
\end{table*}


%% file: tables/inter_table_ablation.tex
\begin{table*}[t!]
\centering
\tabcolsep=0.10cm
\caption{\textbf{Ablation results on inter-dataset.} \textbf{Bold} shows the best result. $\downarrow$ means lower value is better, and vice versa for $\uparrow$.}
\begin{tabular}[ht]{@{}lcccccccc@{}}
\toprule
\multirow{2}{*}{Model} & 
\multirow{2}{*}{\# Params (M)} &
\multirow{2}{*}{BPCER $\downarrow$} &
\multirow{2}{*}{APCER $\downarrow$} &
\multirow{2}{*}{ACER $\downarrow$} &
\multirow{2}{*}{F1 $\uparrow$} &
\multirow{2}{*}{Threshold} &
\multirow{2}{*}{AUC $\uparrow$}\\

                                  &
                                  &                            
                                  &
                                  &                                  
                                  &        \\ 
\midrule
Resnet-18x0.25 ImageNet & 0.71  &  0.4229  & \textbf{0.2360}  & 0.3295  & 0.6366  & 0.9847 & 0.7294\\
Resnet-18x0.25 ImageNet Multihead & 0.72  &  0.4067  & 0.2712  & 0.3390  & 0.6364  & 0.8952 & 0.7145\\
PDGAN Backbone (Ours) & 0.71  &  \textbf{0.3987}  & 0.2469  & \textbf{0.3228}   & \textbf{0.6507}  & 0.9859 & \textbf{0.7326}\\
PDGAN Backbone (Ours) Multihead & 0.72  &  0.4280  & 0.2577  & 0.3429 & 0.6252 & 0.9253 & 0.7121\\

\midrule
\end{tabular}
\vspace{-3mm}
\label{table:inter_ablation}
\end{table*}


%% file: 5_Conclusion.tex
\section{Conclusion}

In this paper, we discovered a new insight for extracting an efficient feature for classifying the spoofed images. We showed that the ability of generating the pre-defined pseudo-depth images for each real and fake images is a key factor for the classification, and proposed a new PDGAN that predicting the pseudo-depth given the input images by catching the details of real and fake images.
We conjectured that the encoder trained to generate features through adversarial training learns important information that is crucial for achieving significant performance, especially on data that it has not seen before.
By the extensive experiments from the public and real-scenario dataset, we showed the validity of our conjectured, and show that the backbone pre-trained by our method outperformed the classification based backbone networks. Also, we achieved the comparable performance to other methods using many auxiliary queues, again supporting our suggestion, the importance of generating the pseudo-depth information.